\documentclass[journal]{IEEEtran}

\ifCLASSINFOpdf
\else
   \usepackage[dvips]{graphicx}
\fi
\usepackage{url}

\usepackage{graphicx}
\usepackage{booktabs}
\usepackage{amsfonts}
\usepackage{hyperref}
\usepackage{caption}
\usepackage{multirow}
\usepackage{float}
\usepackage{subfig}
\usepackage[ruled,linesnumbered]{algorithm2e}
\usepackage{cite}
\usepackage[numbers,sort&compress]{natbib}
\usepackage{amsmath}
\usepackage{cases}
\usepackage{color}
\usepackage{balance}
\usepackage{svg}
\usepackage[normalem]{ulem}
\usepackage[utf8]{inputenc}
\usepackage{url}
\usepackage{amssymb}
\usepackage{bbding}
\usepackage{pifont}
\usepackage{wasysym}
\usepackage{utfsym}
\usepackage{fontawesome}
\usepackage{tabularray}

\begin{document}

\title{AVT$^2$-DWF: Improving Deepfake Detection with Audio-Visual Fusion and Dynamic Weighting Strategies}

\author{Rui Wang, Dengpan Ye \IEEEmembership{Member, IEEE}, Long Tang, Yunming Zhang, Jiacheng Deng 

% \thanks{This paragraph of the first footnote will contain the date on which you submitted your paper for review. It will also contain support information, including sponsor and financial support acknowledgment. For example, ``This work was supported in part by the U.S. Department of Commerce under Grant BS123456.'' }
% \thanks{}
% \thanks{}
}

% \markboth{Journal of \LaTeX\ Class Files, Vol. 14, No. 8, August 2015}
% {Shell \MakeLowercase{\textit{et al.}}: Bare Demo of IEEEtran.cls for IEEE Journals}
% \maketitle

\markboth{2024}
{Wang \MakeLowercase{\textit{et al.}}: Bare Demo of IEEEtran.cls for IEEE Journals}
\maketitle

\begin{abstract}
With the continuous improvements of deepfake methods, forgery messages have transitioned from single-modality to multi-modal fusion, posing new challenges for existing forgery detection algorithms. In this paper, we propose \textbf{AVT$^2$-DWF}, the \textbf{A}udio-\textbf{V}isual dual \textbf{T}ransformers grounded in \textbf{D}ynamic \textbf{W}eight \textbf{F}usion, which aims to amplify both intra- and cross-modal forgery cues, thereby enhancing detection capabilities. AVT$^2$-DWF adopts a dual-stage approach to capture both spatial characteristics and temporal dynamics of facial expressions. This is achieved through a face transformer with an $n$-frame-wise tokenization strategy encoder and an audio transformer encoder. Subsequently, it uses multi-modal conversion with dynamic weight fusion to address the challenge of heterogeneous information fusion between audio and visual modalities. Experiments on DeepfakeTIMIT, FakeAVCeleb, and DFDC datasets indicate that AVT$^2$-DWF achieves state-of-the-art performance intra- and cross-dataset Deepfake detection. Code is available at \url{https://github.com/raining-dev/AVT2-DWF}.
\end{abstract}


\IEEEpeerreviewmaketitle

\section{Introduction}

\IEEEPARstart{W}{ith} the continuous advancement of AI-Generated Content (AIGC) technology, the generation mode is no longer limited to a single modality. Recently, a "HeyGen" tool was utilized to generate a video featuring singer Taylor Swift speaking Chinese, using fabricated lip movements and voice. Such complex and diverse deepfakes pose significant challenges for detection. Therefore, advanced methods are urgently needed to detect these sophisticated deepfake videos.

\begin{figure}[t]
\centering
\includegraphics[width=0.5\textwidth]{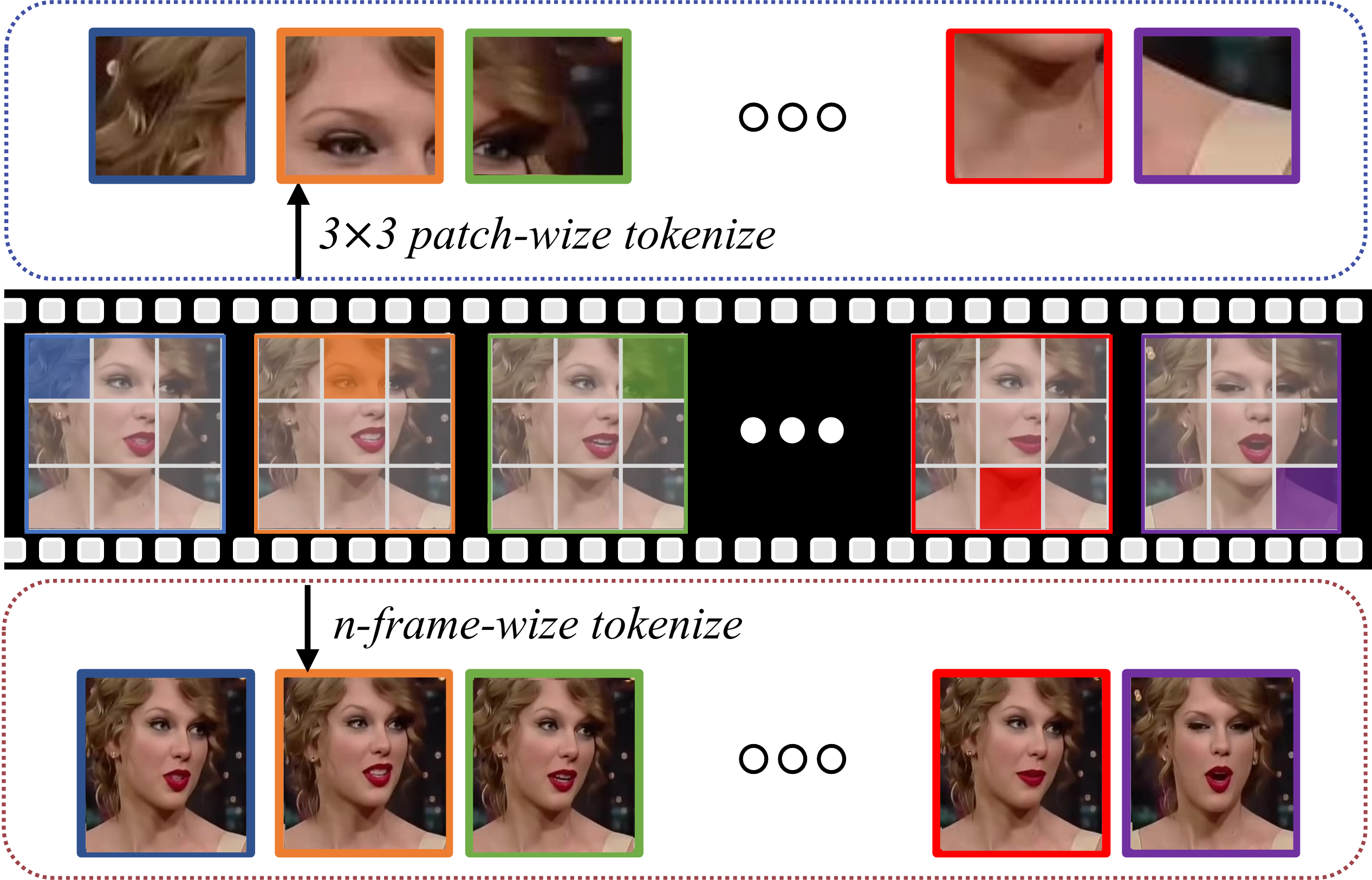}
\caption{\label{fig:fig1}The top image illustrates the conventional approach of packaging video frames into a patch-wise tokenize scheme. The bottom image showcases our proposed method, employing an $n\text{-frame-wise}$ tokenize strategy.}
\end{figure}

Prior methods \cite{verdoliva2020media,rossler2019faceforensics++} mainly focused on single-modal detection, employing established facial manipulation techniques for visual trace recognition and prediction. However, their performance across datasets is subpar. Some existing methods try to utilize patch-level spatiotemporal cues to enhance the robustness and generalization ability of the model \cite{zhang2022deepfake,heo2023deepfake}. These methods build the input video into patch instances processed by a visual transformer, as shown in the top image of Fig.\ref{fig:fig1}. However, this compromises the inherent correlation among facial components, impeding the detection of spatial inconsistencies. Furthermore, audio content can be fabricated, and concentrating exclusively on visual-level authenticity detection will result in bias. Consequently, the domain of multi-modal audio-visual forgery detection has attracted significant attention in research.

Several methods for multi-modal Deepfake detection currently exist. For instance, EmoForen \cite{mittal2020emotions}  focuses on detecting affective inconsistencies, while MDS \cite{chugh2020not} introduces the Modal Discordance Score to quantify audio-visual dissonance. VFD \cite{cheng2023voice} employs a voice-face matching method for forged video detection. AVA-CL \cite{zhang2023joint} leverages audio-visual attention and contrastive learning to enhance the integration and matching of audio and visual features, effectively capturing intrinsic correlations. However, previous research focused too much on the fusion of features between modalities and ignored the optimization of intra-modal feature extraction schemes. To solve this problem, this paper optimizes the extraction of intra-modal features through $n$-frame-patch and uses the DWF module to balance the fusion of cross-modal forgery clues to enhance detection capabilities.

In this work, we propose an Audio-Visual multi-modal Transformer grounded in the Dynamic Weight Fusion principle \textbf{AVT$^2$-DWF}, aiming to capture modality-specific attributes and achieve inter-modal coherence. To enhance the model's representational capabilities and explore spatial and temporal consistency in processed videos, we adopt an $n\text{-frame-wise}$ tokenize strategy focused on facial features within video frames, integrated into the Transformer encoder. A parallel process is applied to the audio domain for feature extraction. To address the imperative need for capturing shared features across distinct modalities, we propose a multi-modal conversion with \textbf{D}ynamic \textbf{W}eight \textbf{F}usion (\textbf{DWF}). This innovative mechanism dynamically predicts audio and video modal weights, facilitating more effective integration of forgery trace and common attribute features, thus enhancing detection capabilities.

In summary, our contributions include:
\begin{itemize}
  \item 
  We employ an $n$-frame-wise tokenization strategy enhancing the extraction of comprehensive facial features within video frames, including subtle nuances of facial expressions, movements, and interactions.
  % We employ an $n$-frame tokenization strategy that enhances the comprehensive extraction of facial features in video frames, including the nuances of facial expressions, actions, and interactions.
  % inter-component relationships.
  \item We propose a multi-modal conversion with Dynamic Weight Fusion (DWF) to enhance the fusion of heterogeneous information from audio and video modalities.
  \item We integrate the above two methods and propose a method termed AVT$^2$-DWF. Through a comprehensive evaluation of widely recognized public benchmarks, we demonstrate the broad applicability and notable effectiveness of AVT$^2$-DWF.
\end{itemize}

\section{Method}
\begin{figure}[t]
\centering
\includegraphics[width=.5\textwidth]{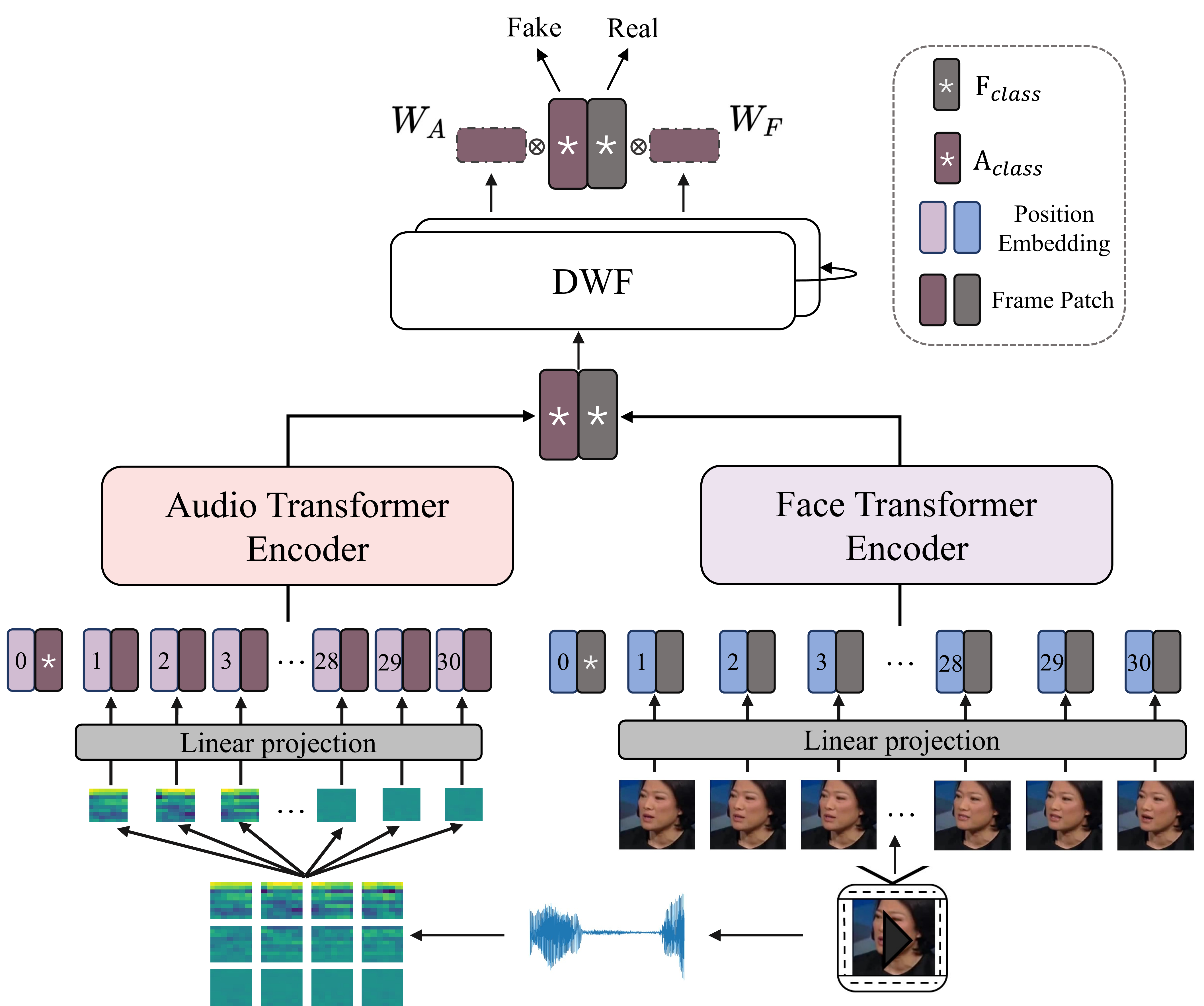}
\caption{\label{fig:fig2}The AVT$^2$-DWF training process is as follows: the audio is combined into MFCC features and fed into the audio conversion encoder for training; at the same time, each group of 30 visual frames is input into the face conversion encoder for training. Their outputs are concatenated and fed into a dynamic weight fusion (DWF) train to obtain audio and visual weight features. These weighted features are multiplied with the outputs of the audio and visual feature encoders and finally concatenated together for detection.
}
\end{figure}

% In response to these challenges, our proposed approach aims to amplify both within-modality and cross-modality forgery cues, thereby enhancing detection capabilities with more practical information. 
Our approach amplifies within-modality and cross-modality forgery cues, enhancing detection capabilities with practical information. Fig.~\ref{fig:fig2} illustrates our proposed AVT$^2$-DWF method, which includes three key components: face transformer encoder, audio transformer encoder, and Dynamic Weight Fusion (DWF) module. First, the face transformer encoder and audio transformer encoder extract visual and audio features to obtain the degree of correlation within the modality. Subsequently, the outputs from both encoders are concatenated and fed into the Dynamic Weight Fusion (DWF) module to train correlation weights between the two modalities, facilitating fusion processing and detection tasks.

\subsection{Face Transformer Encoder}
Face Transformer Encoder stands apart from prior research \cite{zhang2022deepfake,heo2023deepfake} by employing a novel tokenization strategy spanning $n$-frames, as shown in the lower portion of Fig. 1. This strategy redirects the model's focus towards the intrinsic temporal-spatial information across various frames within the video. For a given video $V$, the face block $\mathbf{F} \in \mathbb{R}^{T \times C \times H \times W}$ is extracted. $T$ represents the frame length, $C$ denotes the number of channels, and $H \times W$ corresponds to the frame resolution.  The frames are chronologically reorganized, resulting in a new representation as $C \times (T \times H) \times W$. Similar to the [class] token in ViT \cite{dosovitskiy2020image}, a learnable embedded $\mathbf{F}_{class}$ is incorporated into the sequence, while learnable position embeddings $\mathbf{E}_{p}$ are added. The features of each image patch are linearly mapped to a $D$-dimensional space before entering the Transformer encoder. The Transformer encoder incorporates a multi-head self-attention (MSA) layer, enabling the model to discern correlations among various positions and spatial aspects within the video frame. Layernorm (LN) is applied before every block, and Residual Connections(RC) are applied after every block. The entire process can be formally expressed as:

\begin{align}
    % \begin{aligned}
        \mathbf{F}_0 &= [ \mathbf{F}_{class}\mathbf{E}_{p}; \, \mathbf{f}_1 \mathbf{E}_{p}; \, \mathbf{f}_2 \mathbf{E}_{p}; \cdots; \, \mathbf{f}_T \mathbf{E}_{p} ],   \\
    % \end{aligned} \\
    \mathbf{F}_\ell &= \text{MSA}(\text{LN}(\mathbf{F}_{\ell-1})) + \mathbf{F}_{\ell-1},\quad \ell = 1, \dots, L ,
\end{align}

% &\quad \mathbf{f} \in \mathbb{R}^{(H \times W\times C) \times D}, \, \mathbf{E}_{p} \in \mathbb{R}^{(T + 1) \times D} \\

where $\mathbf{f} \in \mathbb{R}^{(H \times W\times C) \times D}$ represents the visual feature and $\mathbf{E}_{p} \in \mathbb{R}^{(T + 1) \times D}$ is the learnable position embedding.

\subsection{Audio Transformer Encoder}
To handle audio components, a transformer model akin to the face transformer encoder is utilized, capitalizing on its self-attention mechanism to capture internal long-range dependencies within the audio. The study systematically extracts acoustic patterns, temporal dynamics, and other audio-specific features from audio signals. The MFCC feature is computed from the audio signal, yielding components denoted as $\mathbf{A} \in \mathbb{R}^{T \times M}$, where $T$ represents time and $M$ represents frequency elements, which are then linearly projected into a one-dimensional embedding. To capture intrinsic structural correlations from audio spectrograms, a learnable embedded class token $\mathbf{A}_{\text{class}}$ is incorporated into the sequence. Additionally, trainable positional embeddings are introduced. The entire process is delineated in the following formula.

\begin{align}
    \mathbf{A}_0 &= [ \mathbf{A}_{class} \mathbf{E}_{p}; \, \mathbf{a}_1 \mathbf{E}_{p}; \, \mathbf{a}_2 \mathbf{E}_{p}; \cdots; \, \mathbf{a}_T \mathbf{E}_{p} ],  \\
     % \label{eq:embedding} \\
    \mathbf{A}_\ell &= \text{MSA}(\text{LN}(\mathbf{A}_{\ell-1})) + \mathbf{A}_{\ell-1},\quad \ell = 1, \dots, L  .
\end{align}
% &\quad \mathbf{a} \in \mathbb{R}^{(H \times W \times C) \times D}, \, \mathbf{E}_{p} \in \mathbb{R}^{(T + 1) \times D}

where $\mathbf{a} \in \mathbb{R}^{(H \times W \times C) \times D}$ represents the audio feature and $\mathbf{E}_{p} \in \mathbb{R}^{(T + 1) \times D}$ also is the learnable position embedding. The outputs $\mathbf{F}_{class}$ and $\mathbf{A}_{class}$ from the face transformer encoder and audio transformer encoder encompass a variety of in-video information such as visual-spatial details, temporal shifts in audio-visual modalities, and audio content.
\begin{figure}[t]
\centering
\includegraphics[width=.5\textwidth]{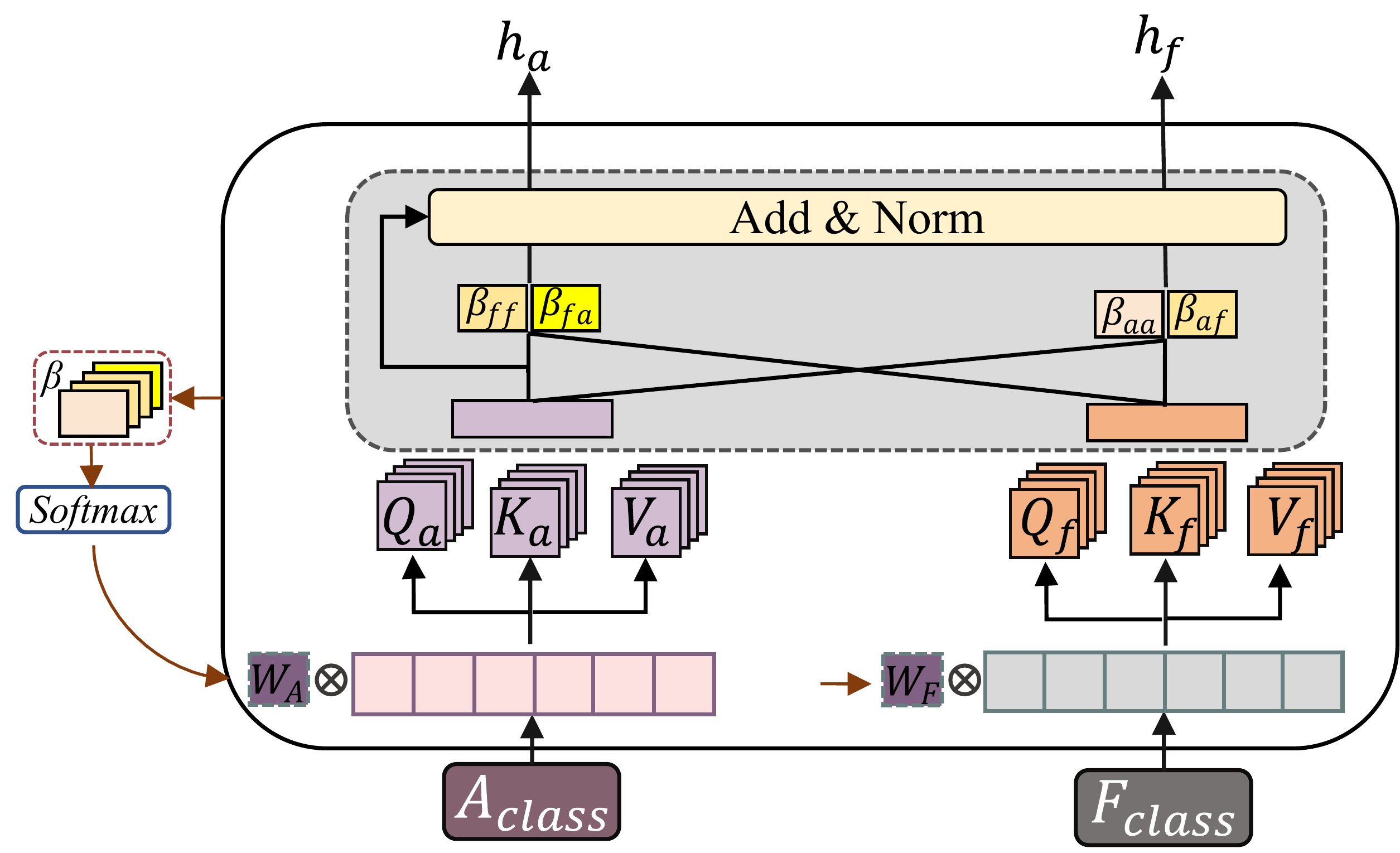}
\caption{\label{fig:fig3} DWF Architecture. The input comprises features $\mathbf{F}_\ell$ and $\mathbf{A}_\ell$ extracted by the face and audio transformer encoders. Initially, weights $W_F$ and $W_A$ are initialized, and the MHCA is utilized to train weight values relevant to the modalities. Subsequently, these weight values are propagated to the subsequent layer of DWF training.}
\end{figure}

\subsection{Multi-Modal Transformer with Dynamic Weight Fusion}
After extracting the audio feature $\mathbf{A}_{class}$ and video feature $\mathbf{F}_{class}$, the DWF module generates entity-level weights $W_A$ and $W_F$ for each modality, as illustrated in Fig. \ref{fig:fig3}. Drawing inspiration from the MEAformer \cite{chen2023meaformer}, our design incorporates a two-layer Multi-Head Cross-modal Attention (MHCA) block to compute these weights. The next layer, MHCA, utilizes the previous layer's weights and does not require initialization. MHCA operates with attention functionality in $N_h$ parallel heads, allowing the model to jointly attend to information from different representation subspaces at different positions. The $i$-th head is parameterized by modal-sharing matrices $W_q^{(i)}$, $W_k^{(i)}$, $W_v^{(i)} \in \mathbb{R}^{d \times d_h}$, which transform the multi-modal input $\mathbf{A}_{class}$, $\mathbf{F}_{class}$ into modality-aware queries $Q_{f/a}^{(i)}$, keys $K_{f/a}^{(i)}$, and values $V_{f/a}^{(i)}$. $d$ represents the dimensionality of the input features, while $d_h$ denotes the dimensionality of the hidden layers. For each feature of modalities, the output is:

\begin{gather}
\text{MHCA}(\mathbf{F}_{class}) = \text{Concat}(W^i_F V_f \cdot W_o), \\
\text{MHCA}(\mathbf{A}_{class}) = \text{Concat}(W^i_A V_a \cdot W_o), \\
W^i_F = \bar{\beta}^{(i)}_{ff} + \bar{\beta}^{(i)}_{fa},  \hspace{0.6cm} W_F= {\textstyle \sum_{i=1}^{N_h}} W_F^i/N_h,\\
W^i_A = \bar{\beta}^{(i)}_{aa} + \bar{\beta}^{(i)}_{af}, \hspace{0.6cm} W_A= {\textstyle \sum_{i=1}^{N_h}} W_A^i/N_h,
\end{gather}

where \(W_o \in \mathbb{R}^{d \times d}\), ${\bar\beta}^{(i)}_{*}$represents the attention weight of the head \(i\). The attention weight of each head ${\bar\beta}^{(i)}_{fa}$ between $f$ and $a$ in each head is defined as follows:

\begin{align}
{\bar\beta}^{(i)}_{fa} = \frac{\exp(Q_f K^{\top}_a / \sqrt{d_h})}  {\textstyle \sum_{n\in {f,a}}\exp(Q_f K^{\top}_n / \sqrt{d_h}) },\quad 
\end{align}
where ${\bar\beta}^{(i)}_{ff}$, ${\bar\beta}^{(i)}_{af}$, and ${\bar\beta}^{(i)}_{aa}$ are similarly calculated, with \(d_h=d/N_h\).LN and RC also stabilize the training.
\begin{align}
h_v=\text{LN} (\text{MHCA}(\mathbf{F}_{\ell-1})+\mathbf{F}_{\ell-1}),\\
h_a=\text{LN} (\text{MHCA}(\mathbf{A}_{\ell-1})+\mathbf{A}_{\ell-1}),
\end{align}

where $h_v$ and $h_a$ are then passed to the next layer of the DWF module for further training.

\textbf{Modal Fusion.} To maximize feature utilization between audio and visual modalities, we multiply previously extracted audio $\mathbf{A}_{class}$, and video features $\mathbf{F}_{class}$ by entity-level weights $W_A$ and $W_F$ in the modal fusion segment. This approach ensures modal diversity and avoids excessive self-focus.
\begin{align}
V = W_F \mathbf{F}_{class}\oplus W_A \mathbf{A}_{class}.
\end{align}

\section{Experiment}
\subsection{Dataset}
The experiments involve three datasets: DeepfakeTIMIT (DF-TIMIT) \cite{korshunov1812deepfakes}, DFDC \cite{dolhansky2020deepfake}, and FakeAVCeleb \cite{khalid2021fakeavceleb}. Since the proportion of real and fake video in these datasets is highly unbalanced, we employ diverse methods to balance real and fake data. Table \ref{tab:tab0} shows the change in the proportion of real and fake data before and after balancing. The raw videos of VidTIMIT \cite{sanderson2002vidtimit} were integrated into the DF-TIMIT dataset. The DFDC dataset extracted partial consecutive frames from each Deepfake video. In contrast, all frames were utilized for training real videos. To address the data imbalance issue in the FakeAVCeleb dataset, 19,000 real videos were selected from VoxCeleb2 \cite{chung2018voxceleb2}. The datasets were partitioned into training, validation, and test sets at a ratio of 7:1:2. The proportion of real and fake data balance in the test set was 1:1. All experimental evaluations were conducted exclusively on the test set.

\begin{table}[]
\centering
\caption{Proportion of Real and Fake Videos Before and After Dataset Balancing, Number of Real: Fake.} \label{tab:tab0}
\scalebox{0.89}{
\begin{tblr}{
  column{1,2,3,4} = {c},
  hline{1,4} = {-}{0.05em},
  hline{2} = {-}{0.05em},
}
\hline
Dataset     & DFDC  & FakeAVCeleb & DF-TIMIT \\
Before & 18:82 & 3:97       & 0:100      \\
After & 43:57 & 44:56    & 50:50       \\
\hline
\end{tblr}}
\end{table}

\subsection{Implementation}
During training, both genuine and synthetic videos are divided into blocks of length $T$ (default is 30). For face detection, the Single Shot Scale-invariant Face Detector (S$^3$FD \cite{zhang2017s3fd}) is employed. Detected faces are then aligned and saved as images with dimensions $224\times224$. In audio processing, MFCC features are computed as input using a 15 ms Hanning window and a 4 ms window shift for accurate spectrum analysis. All experiments were performed under the same settings to ensure the comparability of experimental results.

\begin{table*}[ht]
\centering
\caption{Comparative Analysis of AVT$^2$-DWF Against State-of-the-Art Techniques on DF-TIMIT, FakeAVCeleb, and DFDC Datasets. Evaluation of Detection Performance Using ACC (\%) and AUC (\%). ${\ddag}$: the model is reproduced by ourselves, ——: the authors did not report this metric on this dataset.} \label{tab:tab1}
\scalebox{0.9}{
\begin{tblr}{
  cell{1}{1} = {r=2}{c},
  cell{1}{2} = {r=2}{c},
  cell{1}{3} = {c=2}{c},
  cell{1}{5} = {c=2}{c},
  cell{1}{7} = {c=2}{c},
  cell{1}{9} = {c=2}{c},
  vline{2-4,6,8} = {1}{0.05em},
  vline{5,7,9} = {2}{0.05em},
  vline{2-3,5,7,9} = {-}{0.05em},
  hline{1,14} = {-}{0.05em},
  hline{2} = {3-10}{0.05em},
  hline{3} = {-}{0.05em},
  column{1,2,3,5,7,9} = {c}
}
\hline
Method       & Modality & DF-TIMIT(LQ) &       & DF-TIMIT(HQ) &       & FakeAVCeleb &       & DFDC   &       \\
             &          & ACC          & AUC   & ACC          & AUC   & ACC          & AUC   & ACC    & AUC   \\
Meso-4 \cite{afchar2018mesonet}${\ddag}$             & V        & 49.25        & 50.00    & 51.50         & 50.00    & 59.00           & 60.12 & 50.35  & 50.35 \\
Capsule \cite{nguyen2019use} ${\ddag}$          & V        & 48.25        & 47.99 & 50.25        & 50.98 & 71.43        & 70.41 & 74.21  & 75.62 \\
Xception \cite{rossler2019faceforensics++} ${\ddag}$     & V        & 97.96        & 98.10  & 95.20         & 95.60  & 72.71        & 73.51 & 80.54  & 79.34 \\
Face X-ray \cite{li2020face}   & V        & ——           & 96.95    & ——           & 94.47    & 72.88           & 73.52    & 43.42     & 59.36 \\
CViT \cite{wodajo2021deepfake} ${\ddag}$         & V        & 97.2         & 98.25 & 98.01        & 98.73 & 75.14        & 79.00    & 74.204 & 73.75 \\
LipForensics \cite{haliassos2021lips}${\ddag}$ & V        & 90.75        & 90.71 & 99.25        & 99.27 & 64.00           & 65.23 & 63.60   & 64.50  \\
EmoForen \cite{mittal2020emotions}    & AV       & ——           & 96.30  & ——           & 94.90  & ——           & ——    & ——     & 84.40  \\
MDS \cite{chugh2020not} ${\ddag}$         & AV       & 67.20         & 64.50  & 64.18        & 65.40  & 81.80         & 82.65 & 87.80   & 86.51 \\
VFD \cite{cheng2023voice}         & AV       & ——           & 99.95 & ——           & 99.82 & 81.52        & 86.11 & 80.96  & 85.13 \\
% AVoiD-DF \cite{yang2023avoid}     & AV       & ——           & ——    & ——           & ——    & 83.70         & 89.20  & 91.40   & 94.80  \\
AVA-CL\cite{zhang2023joint}       & AV       & 97.79        & \textbf{99.99} & 96.53        & \textbf{99.86} & 86.55        & \textbf{89.47} & 84.20   & 88.64 \\
AVT$^2$-DWF (ours)          & AV       & \textbf{100.00}        & \textbf{100.00} & \textbf{98.43}        & 98.43 & \textbf{87.57}     & 88.32 & \textbf{88.02}   &  \textbf{89.20}\\ \hline
\end{tblr}}
\end{table*}

\subsection{ Comparisons With The State-of-the-arts}
In comprehensive experiments, the efficacy of AVT$^2$-DWF is evaluated against state-of-the-art baselines using performance metrics such as ACC (Accuracy) and AUC (Area Under the Curve). The baseline models are categorized into two groups: visual modality (V) and multi-modality (AV). A comparative analysis is conducted on three datasets, and the results are presented in Table \ref{tab:tab1}. Most notable outcomes are emphasized in bold, the same hereinafter. Due to the limited quantity of videos, most baseline methods exhibit elevated detection performance on DF-TIMIT. AVT$^2$-DWF and AVA-CL stand out with an accuracy of 99.99\% and 100\% on DF-TIMIT (LQ), significantly surpassing other methods. In the challenging FakeAVCeleb dataset, designed for intricate video forgery, AVA-CL, employing the audio-visual attention contrast learning method, demonstrates comparable performance to our AVT$^2$-DWF. Notably, our approach is more reliable due to a balanced test set. In the expansive DFDC dataset, AVT$^2$-DWF outperforms other vision and audio-visual-based detection methods, achieving an accuracy of 88.02\% and an AUC of 89.20\%, showcasing exceptional performance.

\subsection{Cross-dataset Evaluation}
The robustness assessment of the AVT$^2$-DWF model is prioritized in this phase. To ensure cross-dataset generalizability, our approach is compared with four prominent models: Xception \cite{rossler2019faceforensics++}, CViT \cite{wodajo2021deepfake}, Lipforensis \cite{haliassos2021lips}, and MDS \cite{mittal2020emotions}. The cross-dataset evaluations extend across three benchmark datasets. Specifically, FakeAVCeleb comprises four distinct deep fake methods, DFDC encompasses eight techniques, and DF-TIMIT involves two processes—each dataset presenting unique deepfake challenges. The cross-dataset evaluation results for these three benchmarks are summarized in Table \ref{tab:tab2}. Conventional methods demonstrate subpar performance when confronted with unseen Deepfakes. Although CViT, leveraging transformers as detectors, achieves commendable results, our AVT$^2$-DWF surpasses its performance, demonstrating enhanced efficacy in Deepfake detection.

\begin{table}[tb]
\centering
\caption{AUC (\%) of Cross-datasets Experiments. The Training and Test Sets for Cross-dataset Deepfake Detection are Shown in row 1 and row 2, Respectively.} \label{tab:tab2}
\scalebox{0.76}{
\begin{tblr}{
  column{1,2,3,4,5,6,7} = {c},
  cell{1}{1} = {r=2}{},
  cell{1}{2} = {c=3}{},
  cell{1}{5} = {c=3}{},
  vline{2-3} = {1}{0.05em},
  vline{5} = {2}{0.05em},
  vline{2,5} = {-}{0.05em},
  hline{1,8} = {-}{0.05em},
  hline{2} = {-}{0.05em},
  hline{3} = {-}{0.05em},
}
\hline
Method       & FakeAVCeleb &              &              & DFDC         &              &             \\
             & DFDC         & DF(LQ) & DF(HQ) & FakeAVCeleb & DF(LQ) & DF(HQ) &       \\
Xception \cite{rossler2019faceforensics++}    & 50.00        & 64.06        & 45.01        & 61.43        & 53.75        & 49.79  \\
CViT \cite{wodajo2021deepfake}        & 51.10        & 58.50        & 63.75        & 57.57        & 51.25        & 53.75  \\
LipForensics \cite{haliassos2021lips} & 49.00        & 55.27        & 52.49        & 56.14        & 54.35        & 50.94         \\
MDS \cite{chugh2020not}          & 63.50        & 53.20        & 54.60        & 47.62        & 54.17        & 55.28        \\
AVT$^2$-DWF (ours)         &\textbf{ 74.60}        & \textbf{67.50}        & \textbf{64.1}         & \textbf{77.20 }       & \textbf{66.30}        & \textbf{63.20}   \\  \hline 
\end{tblr}}
\vspace{-0.15cm}
\end{table}

\subsection{Ablation Study}

\subsubsection{Benefit of DWF module}
In a comprehensive evaluation of the AVT2-DWF module, we conducted ablation experiments, examining a purely visual version, an AV version (simply concatenating speech and face extractors), and an AVT2-DWF that combines AV and DWF modules (VA-DWF). The test results on DFDC and FakeAVCeleb datasets are presented in Table \ref{tab:tab3}. In the DFDC dataset, where the audio is not forged, relying solely on spliced audio-visual features for classification leads to a substantial decline in detection results. Conversely, for the FakeAVCeleb dataset, wherein the visual modality of some videos is real while the audio modality is manipulated, the audio-visual module significantly enhances performance. With the introduction of the DWF module, their detection results improved by 11.55\% and 12.89\%, respectively, highlighting the significant advantages of our DWF module in capturing shared features across different modalities.

\begin{table}[tb]
\centering
\caption{AUC (\%) of Detection Results of Integrating Different Modalities on the DFDC and FakeAVCeleb Datasets.}\label{tab:tab3}
\resizebox{.80\columnwidth}{!}{
\begin{tblr}{
  column{1,2,3,4,5} = {c},
  hline{1,5} = {-}{0.05em},
  hline{2} = {-}{0.05em},
}
\hline 
Visual & Audio & VA-DWF
  & DFDC   & FakeAVCeleb    \\
\Checkmark     &       &                       & 85.40 & 70.95 \\
\Checkmark     & \Checkmark     &                       & 77.65 & 75.43  \\
\Checkmark      & \Checkmark     & \Checkmark                     & \textbf{89.20} & \textbf{88.32}  \\ \hline 
\end{tblr}}

\end{table}

\subsubsection{Benefit of $n$-frame-wize tokenize }
To assess the advantages of the $n$-frame-wise tokenization strategy, non-repeating patches are randomly extracted from a sequence of consecutive face frames. These patches are then assembled into complete images for input. The test results in DFDC and FakeAVCeleb are presented in Table \ref{tab:tab4}. On these two benchmarks, the performance of our proposed $n$-frame-wise tokenization strategy improves by 22.45\% and 3.74\%, respectively, compared with the traditional patch method, demonstrating the effectiveness of our system in maintaining continuous information of the entire face. 
% This approach offers significant advantages in capturing spatial inconsistencies in video frames.

\begin{table}[tb]
\centering
\caption{AUC (\%) of Using Two Visual Modality Processing Methods on DFDC and FakeAVCeleb Datasets.}\label{tab:tab4}
\resizebox{.70\columnwidth}{!}{
\begin{tblr}{
  column{1,2,3} = {c},
  vline{2} = {-}{0.05em},
  hline{1,4} = {-}{0.05em},
  hline{2} = {-}{0.05em},
}
\hline 
Visual & DFDC  & FakeAVCeleb \\
\textit{patch-wize tokenize}  & 62.95  & 67.21         \\
\textit{$n$-frame-wize tokenize }  & \textbf{85.40} & \textbf{70.95 } \\ \hline       
\end{tblr}}

\end{table}

\section{Conclusion}
This paper proposes the AVT$^2$-DWF framework to address the subtle spatial variances and temporal consistencies within video content. The unique attributes of each modality are highlighted by using face transformer and audio transformer encoders employing an $n$ frame tokenization strategy. Subsequently, the Dynamically Weighted Fusion (DWF) technique extracts common attributes from the audiovisual modalities. Our experimental results indicate superior performance of AVT$^2$-DWF in both intra- and cross-dataset executions compared to other Deepfake detection methods. These findings suggest that achieving ubiquitous consistency across multiple modalities can effectively serve as a critical indicator for Deepfake detection in real-world scenarios.
% \section*{References}
% \subsection*{Examples:}
% \def\refname{}
% \begin{thebibliography}{34}
\bibliographystyle{IEEEtran}
\balance

\end{document}